\documentclass[10pt,twocolumn,letterpaper]{article}

\usepackage[pagenumbers]{iccv} 

\definecolor{iccvblue}{rgb}{0.21,0.49,0.74}
\usepackage[pagebackref,breaklinks,colorlinks,allcolors=iccvblue]{hyperref}

\title{SARA: Structural and Adversarial Representation Alignment for
Training-efficient Diffusion Models}

\author{
    Hesen Chen$^{2}$ \quad 
    Junyan Wang$^{3}$ \quad 
    Zhiyu Tan$^{1,2}$ \quad 
    Hao Li$^{1,2}$\textsuperscript{†} 
    \vspace{2mm}\\
    $^1$Fudan University \\
    $^2$Shanghai Academy of Artificial Intelligence for Science \\
    $^3$Australian Institute for Machine Learning, The University of Adelaide \\
}

\begin{document}
\maketitle
\begin{abstract}
Modern diffusion models encounter a fundamental trade-off between training efficiency and generation quality. While existing representation alignment methods, such as REPA, accelerate convergence through patch-wise alignment, they often fail to capture structural relationships within visual representations and ensure global distribution consistency between pretrained encoders and denoising networks. To address these limitations, we introduce SARA, a hierarchical alignment framework that enforces multi-level representation constraints: (1) patch-wise alignment to preserve local semantic details, (2) autocorrelation matrix alignment to maintain structural consistency within representations, and (3) adversarial distribution alignment to mitigate global representation discrepancies. Unlike previous approaches, SARA explicitly models both intra-representation correlations via self-similarity matrices and inter-distribution coherence via adversarial alignment, enabling comprehensive alignment across local and global scales. Experiments on ImageNet-256 show that SARA achieves an FID of 1.36 while converging twice as fast as REPA, surpassing recent state-of-the-art image generation methods. This work establishes a systematic paradigm for optimizing diffusion training through hierarchical representation alignment.
\end{abstract}    
\section{Introduction}
\label{sec:intro}

\begin{figure}[t]
  \centering
    \includegraphics[width=0.47\textwidth]{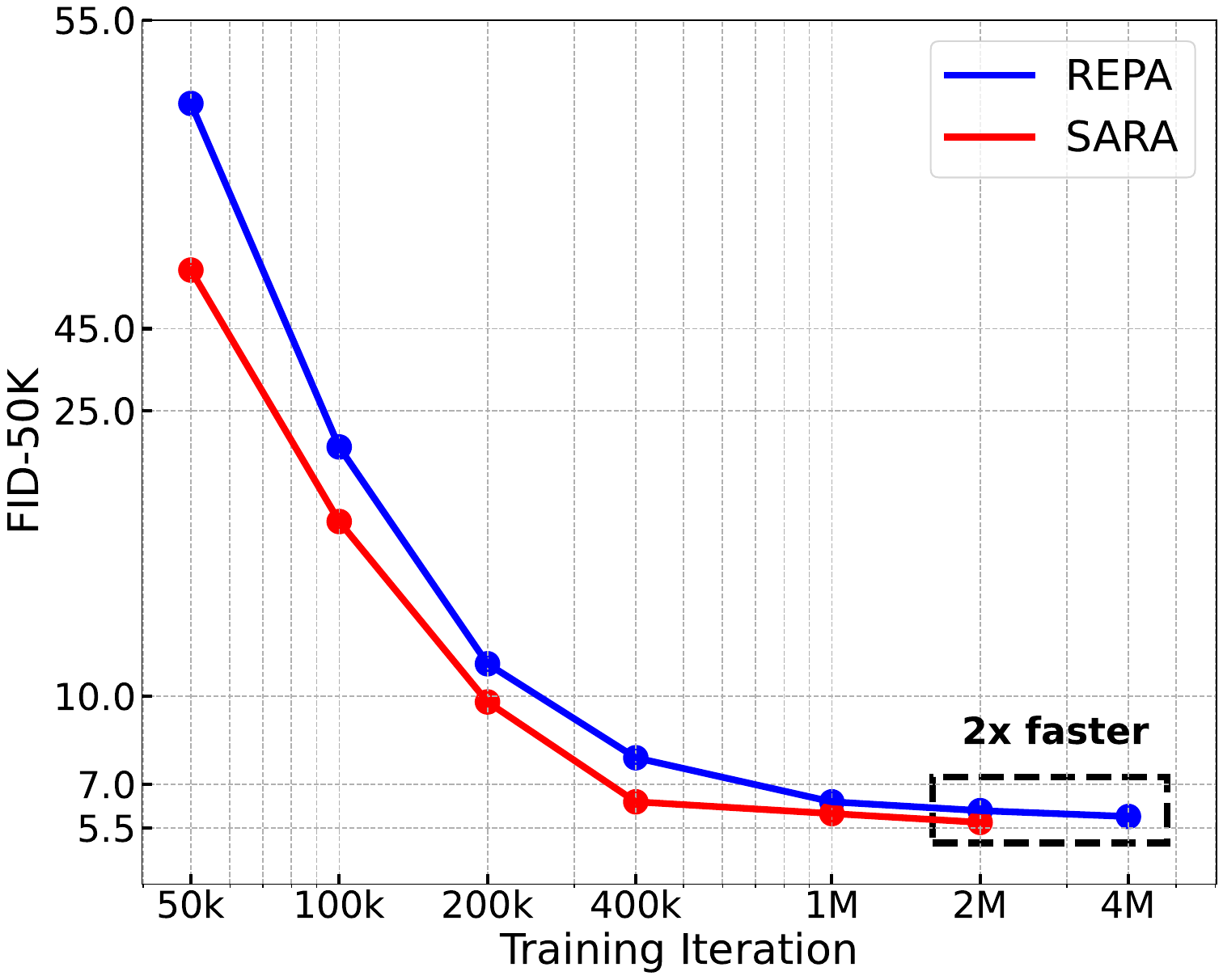}
   \caption{SARA enhances the efficiency and effectiveness of diffusion model training through multi-level representation alignment. Compared to REPA, it achieves 2$\times$ faster convergence speed, significantly accelerating the training process.}
   \label{fig:main}
\end{figure}

Denoising-based generative models, including diffusion \cite{ho2020denoising,songdenoising,songscore} and flow-based approaches \cite{liuflow,lipman2022flow,esser2024scaling}, have emerged as powerful and scalable methods for generating high-dimensional visual data. These models have achieved remarkable success, particularly in complex tasks such as zero-shot text-to-image \cite{dhariwal2021diffusion,rombach2022high} and text-to-video \cite{yang2024cogvideox,blattmann2023stable} generation. Recent advances in vision generation emphasize the importance of effective representation alignment between pre-trained visual encoders \cite{oquab2024dinov2,chen2021empirical,he2022masked} and denoising networks \cite{ma2024sit,peebles2023scalable}. The REPA framework \cite{yu2024representation} introduced a patch-wise alignment strategy to bridge the gap between these representation spaces. However, despite its advantages, patch-wise alignment suffers from two fundamental limitations:

\noindent
\textbf{Structural Fragmentation}: Treating representations as isolated patches disregards spatial and semantic relationships between visual elements, such as part-whole hierarchies in objects.

\noindent
\textbf{Global Distribution Misalignment}: Local representation matching alone fails to enforce global consistency, leading to internal information inconsistencies that degrade representation quality and ultimately harm image generation.

These issues stem from an oversimplified alignment strategy that overlooks the multi-scale nature of visual representations, particularly in large-scale diffusion frameworks and complex generation tasks. To address these challenges, we propose a more comprehensive alignment approach operating at three complementary levels:

\noindent
\textbf{Local}: Ensuring patch-wise semantic fidelity by aligning individual representations while preserving their local semantic integrity.

\noindent
\textbf{Structural}: Capturing and maintaining relational consistency within representations to reflect structural relationships.

\noindent
\textbf{Global}: Aligning overall representation distributions to ensure consistency across models.

Building on these three complementary alignment levels, we introduce SARA (Structural and Adversarial Representation Alignment), a unified framework that extends REPA by integrating two key innovations to enhance representation alignment.

First, we propose structured autocorrelation alignment, which computes the autocorrelation of representations separately from both pre-trained visual encoders and denoising networks, then aligns their autocorrelation matrices. Unlike the patch-wise alignment in REPA, this method leverages structural relationships within the representation space, providing a more holistic alignment strategy that preserves internal correlations.

Second, to refine the global representation distribution, we introduce a GAN-based loss \cite{goodfellow2020generative}, which enhances alignment between visual encoders and denoising networks. Acting as a regularizer, the adversarial loss encourages the denoising network to produce representations that better match the global distribution of pre-trained encoders, ultimately improving representation fidelity and generation quality.

By integrating these innovations with the original patch-wise alignment from REPA, SARA establishes a robust alignment framework that significantly improves both training efficiency and final image quality.

The main contributions of this paper can be summarized as follows:
\begin{itemize}
    \item Building on the assumptions of REPA, we propose that high-quality representations in diffusion transformers should be learned by considering both structural integrity and holistic distribution consistency.
    \item We introduce structural autocorrelation alignment, a technique that enhances the structural coherence of denoising network representations.
    \item We propose adversarial distribution alignment, which refines the global representation matching between visual encoders and denoising networks.
    \item We present \textbf{SARA}, a unified framework that integrates patch-wise alignment, structural autocorrelation alignment, and adversarial distribution alignment.
\end{itemize}

Extensive experiments demonstrate that SARA achieves state-of-the-art performance on standard image generation benchmarks, significantly surpassing existing methods and offering new insights into the importance of representation alignment in generative models.

\begin{figure*}
  \centering
  \begin{subfigure}{0.6\linewidth}
    \includegraphics[width=\textwidth]{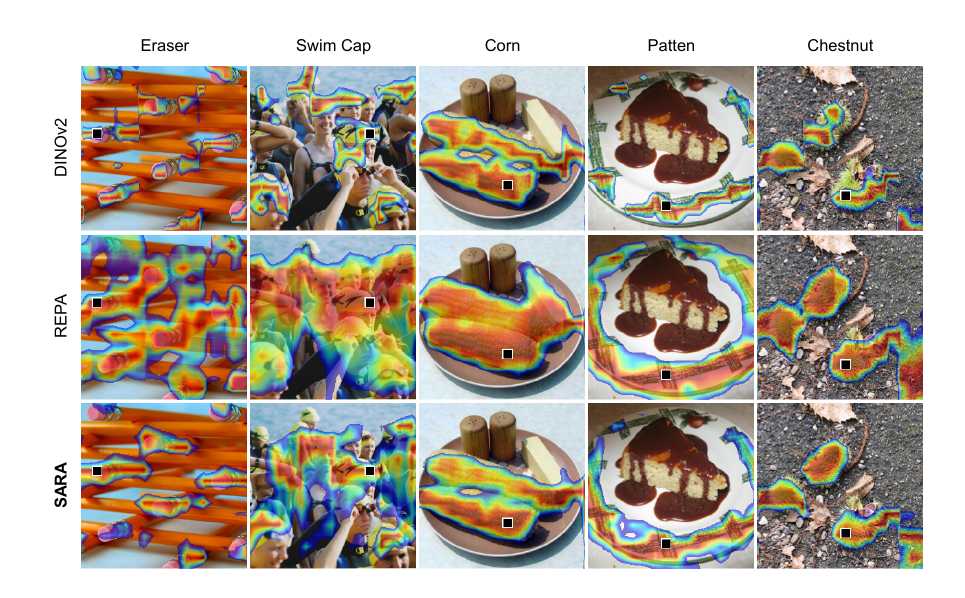}
    \caption{Visualization of Representation Correlations. Heatmaps illustrating the correlation of a selected patch (marked with a black square) within the intermediate representations of DINOv2, REPA, and SARA.}
    \label{fig:heatmap}
  \end{subfigure}
  \hfill
  \begin{subfigure}{0.36\linewidth}
    \includegraphics[width=\textwidth]{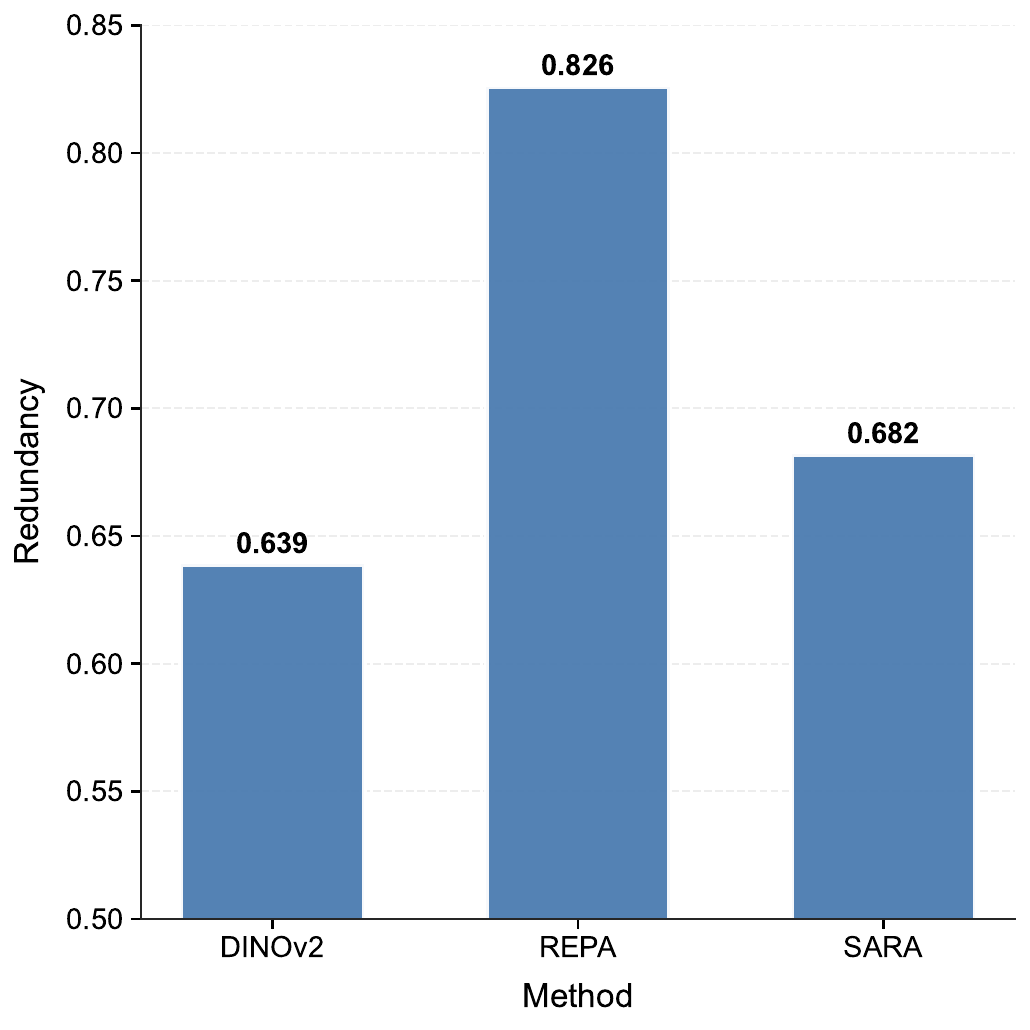}
    \caption{Representation redundancy analysis. The redundancy measured by the energy percentage of the top 50 singular values from representation.}
    \label{fig:redundancy}
  \end{subfigure}
  \caption{Analysis of representation correlations and redundancy across DINOv2, REPA, and SARA.}
  \label{fig:analy}
\end{figure*}
\section{Related Work}
\label{sec:related}

\textbf{Visual Generation with Diffusion Models.} Diffusion models have emerged as a dominant approach in generative computer vision, achieving remarkable success in generating high-quality images and videos \cite{ho2020denoising,songdenoising,songscore,rombach2022high,dhariwal2021diffusion,yang2024cogvideox,blattmann2023stable}. Recent research has focused on enhancing both the training efficiency and inference speed of these models. One major direction is model distillation \cite{salimansprogressive,meng2023distillation}, which reduces the number of sampling steps, accelerating inference without sacrificing quality. Consistency models \cite{song2023consistency,luo2023latent} further improve efficiency by enforcing consistency along the ODE trajectory, leading to a more stable generation process with lower computational cost. Additionally, the integration of transformers into diffusion models has led to notable performance gains \cite{bao2023all,gao2023masked,hatamizadeh2024diffit}. In particular, DiT \cite{peebles2023scalable} and SiT \cite{ma2024sit} have become foundational architectures for state-of-the-art text-to-image and text-to-video generation. Our work builds upon these architectures by introducing multi-level representation alignment.

\noindent
\textbf{Representation learning for diffusion models.} Improving learned representations in diffusion models has been a central focus of recent research \cite{fuest2024diffusion}. Some approaches explore hybrid diffusion models, which jointly perform generation alongside tasks such as classification, detection, and segmentation \cite{yang2022your,deja2023learning,tian2023addp}. Additionally, various knowledge distillation strategies have been developed to transfer information from diffusion models to downstream tasks, improving generalization and efficiency \cite{yang2023diffusion,li2023dreamteacher}. A particularly promising direction is leveraging pre-trained encoders to guide the training of denoising diffusion models, a strategy that has proven essential for enhancing efficiency \cite{pernias2023wurstchen,li2025return,yu2024representation}. Models such as Würstchen \cite{pernias2023wurstchen} and RGC \cite{li2025return} adopt a two-stage diffusion framework, where the first-stage model generates latent representations (e.g., semantic maps or latent vectors), which the second-stage model then uses to synthesize the final images. REPA \cite{yu2024representation} pioneered patch-wise alignment, demonstrating that leveraging semantic-rich representations from pre-trained visual encoders can effectively guide the denoising process and accelerate training. Building on REPA, we introduce multi-level alignment for a more structured use of pretrained representations, improving efficiency and generation quality.

\noindent
\textbf{Adversarial learning for or diffusion models.} Generative adversarial networks (GANs) have been extensively studied in image generation \cite{goodfellow2020generative,karras2017progressive,johnson2016perceptual,liu2018auto,reed2016generative}. Adversarial learning is formulated as a minimax optimization between a discriminator and a generator, where the discriminator learns to distinguish real from generated samples, encouraging the generator to align its outputs with the real data distribution. Recent works have explored incorporating adversarial learning into diffusion models, particularly for training \cite{xiao2021tackling} and distillation \cite{xu2024ufogen,sauer2024adversarial}. For instance, SDXL-Turbo \cite{sauer2024adversarial} employs a pre-trained image encoder as a discriminator backbone to accelerate training, while SDXL-Lightning \cite{lin2024sdxl} further refines this approach by using the diffusion model’s U-Net encoder as the discriminator, enabling effective distillation in the latent space of high-resolution models. In contrast, our method introduces a lightweight discriminator that directly enforces adversarial learning at the intermediate-layer representations of both the pre-trained visual encoder and the denoising generator.

\begin{figure*}[t]
  \centering
  \includegraphics[width=0.95\textwidth]{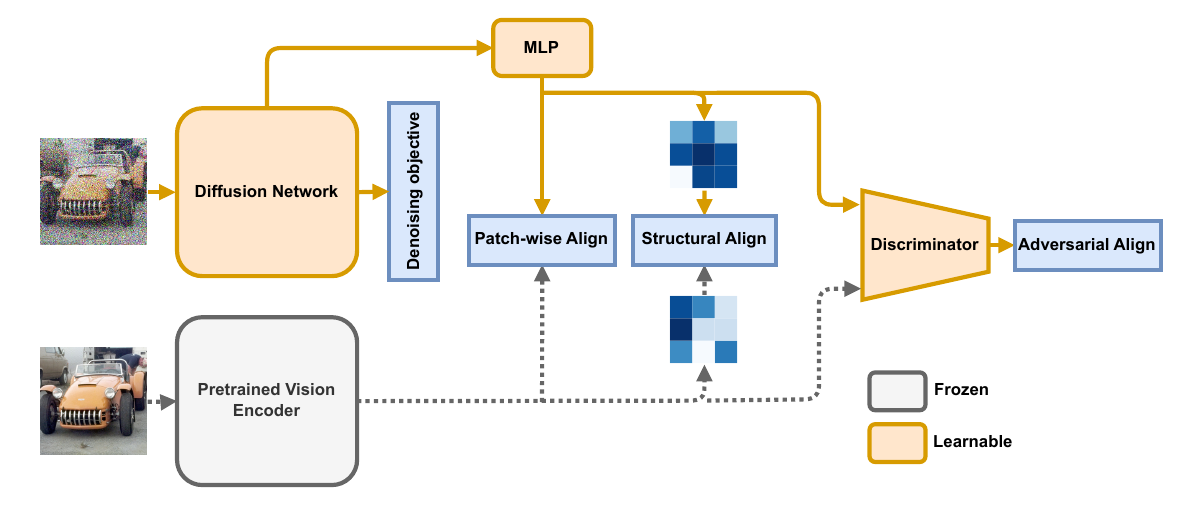}
  \caption{Overview of the SARA framework. SARA aligns diffusion model representations with powerful pretrained visual features through a combination of complementary alignment strategies.}
  \label{fig:framework}
\end{figure*}
\section{Method}
\label{sec:method}

Our approach begins with a thorough analysis of the limitations of REPA’s patch-wise alignment strategy. Based on these insights, we propose SARA, a unified framework that integrates patch-wise alignment, structural autocorrelation alignment, and adversarial distribution alignment. We first provide a comprehensive overview of the SARA framework, detailing its overall architecture. We then present the specific implementations of the three alignment strategies, highlighting their individual contributions. Finally, we introduce a unified training strategy that seamlessly integrates these alignment techniques, ensuring efficient and effective representation learning.

\subsection{Empirical Analysis}
To systematically diagnose the shortcomings of REPA’s patch-wise alignment strategy, we conduct two interconnected analyses that jointly reveal its structural and distributional inconsistencies. These experiments leverage intermediate representations from the DINOv2 ViT-L/14 encoder \cite{oquab2024dinov2}, the REPA-trained diffusion model, and our SARA framework, all operating on identical patch-wise representation dimensions.

We first analyze the spatial-semantic relationships encoded in intermediate representations by visualizing their patch-wise autocorrelation patterns. Specifically, we select a set of images and extract intermediate representations using DINOv2, REPA, and SARA, respectively. For each representation, we compute the correlation between each patch and all other patches. In each image, we designate a specific reference patch (marked with a black square) and visualize its correlation matrix by projecting it onto the corresponding locations in the original image as a heatmap. As shown in Figure \ref{fig:heatmap}, DINOv2’s representations exhibit strong, localized correlations—for instance, in the second column, a black patch located in the swim cap region shows high correlations with other semantically related patches (e.g., other regions of the swim cap), forming a coherent semantic structure. In stark contrast, REPA’s correlations appear scattered and weak. The same black patch fails to highlight semantically related regions, indicating that patch-wise alignment disrupts the intrinsic spatial hierarchies critical for accurate representation learning. In comparison, SARA significantly enhances these correlations, yielding intermediate representations that better preserve spatial-semantic relationships, aligning more closely with those of DINOv2.

Given the complexity of evaluating the consistency of global representation distribution, we employ singular value decomposition (SVD) to measure representation redundancy. By analyzing the energy distribution of representations extracted from 50,000 images, we assess how effectively the representation space captures semantic diversity. As shown in Figure \ref{fig:redundancy}, DINOv2 exhibits a well-balanced energy distribution across dimensions, indicating a rich and diverse high-dimensional representation. However, REPA disrupts this diversity, with its top 50 singular values accounting for 82.6\% of the total energy. This suggests severe redundancy, implying that REPA’s reresentations collapse into a narrow subspace, losing the global distribution properties of the pre-trained representations. In contrast, SARA preserves a more balanced energy distribution, closely aligning with DINOv2. This demonstrates that SARA effectively maintains the diversity and expressiveness of the learned representations, ensuring a more faithful preservation of the original representation space.

\begin{figure*}
  \centering
  \includegraphics[width=\textwidth]{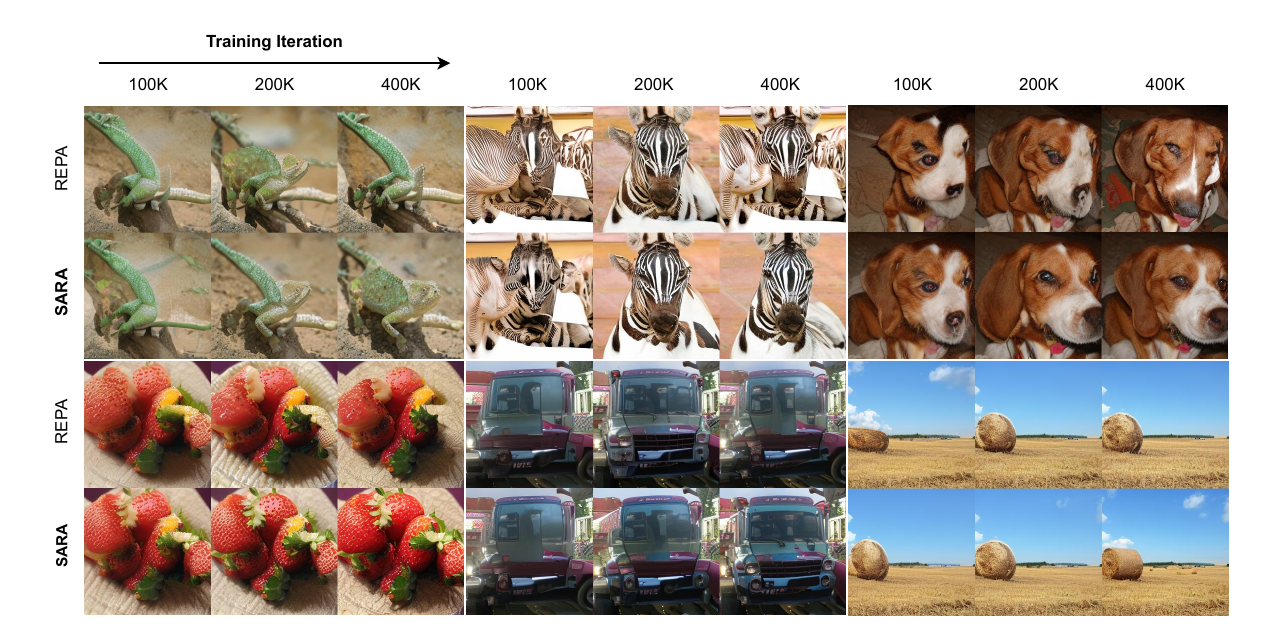}
  \caption{Generated Image Comparison of REPA vs SARA. We compare images generated by two SiT-XL/2 models over the first 400K iterations, with one model using REPA and the other using SARA. Both models are initialized with the same noise, employ the same sampler and number of sampling steps, and do not use classifier-free guidance.}
  \label{fig:iteration}
\end{figure*}

\subsection{Overview}
REPA has demonstrated that reducing the gap between diffusion model representations and pretrained visual model representations can significantly improve both training efficiency and generation quality. However, our analysis reveals that REPA’s representations still exhibit substantial discrepancies at a deeper level. To address this, we propose SARA (Structural and Adversarial Representation Alignment)—a multi-level representation alignment framework that builds upon REPA to further bridge this gap and enhance representation consistency.

As illustrated in Figure \ref{fig:framework}, SARA is a unified framework designed to improve representation alignment in diffusion models. It integrates three key strategies: patch-wise alignment, structural autocorrelation alignment, and adversarial distribution alignment. By leveraging multi-level alignment, SARA ensures that the learned representations in the denoising diffusion model align more effectively with those of a pretrained vision encoder, improving both training efficiency and generation quality. Pretrained vision encoder, Diffusion network, MLP projection layer and Discriminator. Each component plays a crucial role in ensuring robust representation alignment and high-quality image generation.

\noindent
\textbf{Pretrained Vision Encoder.} A frozen visual encoder extracts representations from real images, serving as a strong prior derived from natural images. This ensures that the extracted representations retain meaningful semantics. Given a clean image $X_0$, the pretrained visual encoder $E_\phi$ computes its representations as follows:
\begin{equation}
    \label{eq1}
    z_{enc}=E_\phi(X_0) \in \mathbb{R}^{N \times D}
\end{equation}
where $N, D > 0$ are the number of patches and the embedding dimension of
$E_\phi$. The extracted representation $z_{enc}$ serve as the target for alignment.

\noindent
\textbf{Diffusion Network.} A transformer-based network $f_\theta$ refines noisy latent representations into high-quality image representations. Similar to REPA, our diffusion network is trained with a velocity prediction objective to enhance the generative process. Given a noisy latent representation $x_t$ at timestep $t$, the goal is to predict the velocity of the clean data $v$, defined as:
\begin{equation}
    \label{eq2}
    v_t=\frac{x_t-x_0}{\sigma_t}
\end{equation}
where $x_0$ is the clean latent representation, and $\sigma_t$ is the noise level at timestep $t$. The diffusion network is optimized to minimize the velocity prediction loss:
\begin{equation}
    \label{eq3}
    {\cal{L}}_{velocity}(\theta)=\mathbb{E}_{t,x_0,\epsilon}[||f_\theta(x_t,t) - v_t)||_2^2]
\end{equation}
where $\epsilon\sim{\cal{N}}(0,I)$ represents Gaussian noise. This objective ensures that the diffusion network learns an effective generative process for reconstructing clean image representations from noisy inputs.

Additionally, another learning objective of this module is to enable the diffusion transformer’s hidden states to predict the pretrained clean visual representation from a noise-corrupted input while preserving useful semantic information. Given a noisy input image $X_t$, the diffusion network extracts intermediate representations, which are then utilized for subsequent multi-level alignment:
\begin{equation}
    \label{eq4}
    z^{(i)}_{den}=f^{(i)}_\theta(X_t)
\end{equation}
where $z^{(i)}_{den}$ represents the learned representation of $i_{th}$ layer of diffusion network. 

\noindent
\textbf{MLP Projection Layer.} A multilayer perceptron $M_{proj}$ that maps extracted representations to a shared latent space where alignment occurs. This module ensures that representations from different domains are comparable:
\begin{equation}
    \label{eq5}
    h^{(i)}_{den}=M_{proj}(z^{(i)}_{den}) \in \mathbb{R}^{N \times D}
\end{equation}
where $h^{(i)}_{den}$ is the projected representation in a shared space.

\noindent
\textbf{Discriminator}
An adversarial discriminator that helps align global representation distributions by distinguishing between representations extracted from pretrained vision encoder and those generated by the diffusion network. Since adversarial training operates at the representation level, we employ a lightweight pretrained CNN as the discriminator to reduce computational overhead.

\begin{table*}[ht]
\begin{subtable}{0.35\textwidth}
    \centering
    \resizebox{!}{0.15\textheight}{
    \begin{tabular}{lc@{\hspace{50pt}}c@{\hspace{50pt}}cc}
    \toprule
    Model & \#Params & Iter. & FID↓  \\
    \midrule
    DiT-L/2 & 458M & 400K & 23.3  \\
    + REPA & 458M & 400K & 15.6   \\
    \textbf{+ SARA} & \textbf{458M} & \textbf{400K} & \textbf{13.5}  \\
    \midrule
    DiT-XL/2 & 675M & 400K & 19.5  \\
    + REPA & 675M & 400K & 12.3   \\
    \textbf{+ SARA} & \textbf{675M} & \textbf{400K} & \textbf{10.3}  \\
    \midrule
    SiT-B/2 & 130M & 400K & 33.0 \\
    + REPA & 130M & 400K & 24.4 \\
    \textbf{+ SARA} & \textbf{130M} & \textbf{400K} & \textbf{22.1} \\
    \midrule
    SiT-L/2 & 458M & 400K & 18.8  \\
    + REPA & 458M & 400K & 10.0  \\
    \textbf{+ SARA} & \textbf{458M} & \textbf{400K} & \textbf{8.5} \\
    \midrule
    SiT-XL/2 & 675M & 7M & 8.3  \\
    + REPA & 675M & 400K & 7.9  \\
    \textbf{+ SARA} & \textbf{675M} & \textbf{400K} & \textbf{6.4} \\
    + REPA & 675M & 2M & 6.1 \\
    \textbf{+ SARA} & \textbf{675M} & \textbf{1M} & \textbf{6.0} \\
    + REPA & 675M & 4M & 5.9 \\
    \textbf{+ SARA} & \textbf{675M} & \textbf{2M} & \textbf{5.7} \\
    \bottomrule
    \end{tabular}}
    \captionsetup{width=0.95\textwidth}
    \caption{FID comparisons with DiTs, SiTs, and REPA without classifier-free guidance (CFG). Iter. denotes the training iteration.}
    \label{tab:wocfg}
\end{subtable}
\begin{subtable}{0.64\textwidth}
    \centering
    \resizebox{!}{0.15\textheight}{
    \begin{tabular}{lc@{\hspace{30pt}}c@{\hspace{30pt}}c@{\hspace{30pt}}c@{\hspace{30pt}}c@{\hspace{30pt}}c@{\hspace{30pt}}cc}
    \toprule
    Model & Epochs & FID↓ & sFID↓ & IS↑ &Prec.↑ &Rec.↑ \\
    \midrule
    ADM-U & 400  & 3.94 & 6.14 & 186.7 & 0.82 & 0.52 \\
    VDM++ & 560  & 2.40 & - & 225.3 & -  & - \\
    Simple diffusion & 800  & 2.77 & - & 211.8 &  - & - \\
    CDM & 2160 & 4.88 & - & 158.7 & - & - \\
    \midrule
    LDM-4 & 200 & 3.6 & - & 247.7 & 0.87 & 0.48 \\
    \midrule
    U-ViT-H/2 & 240 & 2.29 & 5.68 & 263.9 & 0.82 & 0.57 \\
    DiffiT* & - & 1.73 & - & 276.5 & 0.80 & 0.62 \\
    MDTv2-XL/2* & 1080 & 1.58 & 4.52 & 314.7 & 0.79 & 0.65 \\
    \midrule
    MaskDiT & 1600 & 2.28 & 5.67 & 276.6 & 0.80 & 0.61 \\
    SD-DiT & 480 & 3.23 & - & - & - & -  \\
    \midrule
    DiT-XL/2 & 1400 & 2.27 & 4.60 & 278.2 & 0.83 &0.57   \\ 
    \midrule
    SiT-XL/2 & 1400 & 2.06 & 4.50 & 270.3 & 0.82 & 0.59 \\
    + REPA & 200  & 1.96 & \textbf{4.49} & 264.0 & 0.82 & 0.60 \\
    \textbf{+ SARA} & \textbf{200} & \textbf{1.82} & 4.51 & \textbf{279.8} & \textbf{0.83} & \textbf{0.60} \\
    \midrule[0.01pt]
    + REPA & 800  & 1.80 & 4.50 & 284.0 & 0.81 & 0.61 \\
    \textbf{+ SARA} & \textbf{400} & \textbf{1.75} & \textbf{4.48} & \textbf{288.9} & \textbf{0.82} & \textbf{0.61} \\
    \midrule[0.01pt]
    + REPA* & 800  & 1.42 & 4.70 & 305.7 & 0.80 & \textbf{0.65} \\    
    \textbf{+ SARA*} & \textbf{400} & \textbf{1.36} & \textbf{4.63} &  \textbf{316.7} &  \textbf{0.81} & 0.63 \\
    \bottomrule
    \end{tabular}}
    \captionsetup{width=0.9\textwidth}
    \caption{Evaluate on ImageNet 256×256 with classifier-free guidance (CFG). Results that include additional CFG scheduling are marked with an asterisk (*), where the guidance interval from is applied for REPA and SARA.}
    \label{tab:cfg}
\end{subtable}
\caption{Quantitative comparison of SARA, REPA, and other diffusion models on ImageNet 256×256.  ↓ and ↑ indicate whether lower or higher values are preferable, respectively.}
\label{tab:1}
\end{table*}

\subsection{Patch-wise Alignment}
The first strategy in SARA is aligning representation at the individual patch or spatial location level like REPA. In particular , it achieves alignment through a maximization of patch-wise similarities between
the pretrained representation $z_{enc}$ and the hidden state $h_{den}$:
\begin{equation}
    \label{eq6}
    {\cal{L}}_{patch}(\theta, {proj})=-\mathbb{E}_{t,x_0,\epsilon}[\frac{1}{N}\sum_{n=1}^Nsim(z^{[n]}_{enc}, h^{[n]}_{den})]
\end{equation}
where $n$ is a patch index and $sim({\cdot}, {\cdot})$ is a pre-defined similarity function. This term forces the generated representations to match the real  representations directly, ensuring local consistency in the learned representation maps. However, while effective, this alignment only enforces similarity at each individual spatial location without considering the broader structure and relationships within the representations maps. This limitation motivates our next alignment method.

\subsection{Structural Alignment}
Patch-wise alignment does not explicitly enforce consistency in the structural relationships between representations. To address this, we introduce structural alignment, which captures the internal coherence of representation representations. For a given hidden representation $h$, we define its autocorrelation matrix $A(h)$ as:
\begin{equation}
    \label{eq7}
    A(h)_{i,j}=\frac{h_i{\cdot}h_j}{\Vert h_i \Vert \Vert h_j \Vert }
\end{equation}
where $h_i$ and $h_j$ are representation vectors at different spatial locations, and $\cdot$ denotes the dot product. This matrix captures how different representation locations interact, providing a measure of internal structure. We enforce structural consistency by minimizing the discrepancy between the autocorrelation matrices of the pretrained visual encoder and the diffusion network:
\begin{equation}
    \label{eq8}
    {\cal{L}}_{struc} = \Vert A(z_{enc})-A(h_{den}) \Vert_{F}^2
\end{equation}
where $\Vert {\cdot} \Vert_{F}$ denotes the Frobenius norm. This encourages the diffusion network to maintain the same internal relationships as the pretrained visual encoder, preserving the structure of natural image representations.

\subsection{Adversarial Alignment}
To ensure global consistency in representation distributions, we employ a discriminator $D_{\psi}$ to distinguish between pretrained vision representations $z_{enc}$ and diffusion representations $h_{den}$. The discriminator is optimized with the following loss:
\begin{equation}
    \label{eq9}
    {\cal{L}}_{D}=-\mathbb{E}[\log{D_\psi(z_{enc})}]-\mathbb{E}[\log(1-D_\psi(h_{den}))]
\end{equation}
In response, the diffusion network is trained to generate representations that are indistinguishable from real ones:
\begin{equation}
    \label{eq10}
    {\cal{L}}_{adv}=-\mathbb{E}[\log{D_\psi(h_{den})}]
\end{equation}

\subsection{Joint Training and Optimization}
Our final training objective combines all alignment losses and denoising loss:
\begin{equation}
    \label{eq11}
    {\cal{L}}={\cal{L}}_{velocity}+\lambda{\cal{L}}_{patch}+\beta{\cal{L}}_{struc}+\gamma{\cal{L}}_{adv}
\end{equation}
where $\lambda$, $\beta$, $\gamma$ are weighting coefficients for different loss terms. By integrating patch, structural, and distributional alignment, our method producing results with higher consistency and realism than previous approaches.
\section{Experiments}
\label{sec:exp}

\begin{figure*}
  \centering
    \includegraphics[width=\textwidth]{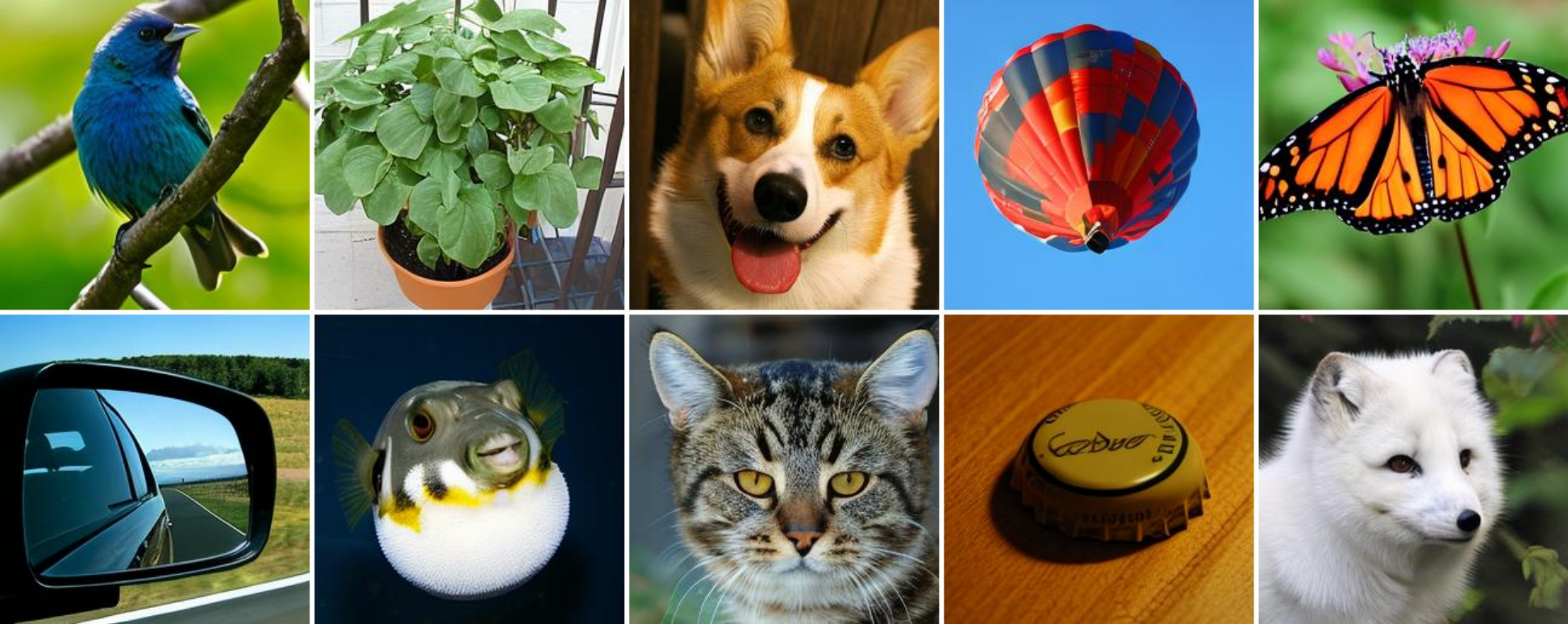}
  \caption{Selected samples from the SiT-XL/2+SARA model on ImageNet 256×256, generated using classifier-free guidance with $w=4.0$.}
  \label{fig:select}
\end{figure*}

We conduct extensive experiments to evaluate the performance of SARA and assess the impact of the proposed components. Specifically, Section \ref{subsec:setup} details the experimental setup, Section \ref{subsec:result} presents the main results, and Section \ref{subsec:ablation} and \ref{subsec:discuss} provide an in-depth ablation study and discussion.

\subsection{Setup}
\label{subsec:setup}
\textbf{Implementation details.} Our experimental setup closely follows REPA \cite{yu2024representation}, which builds upon DiT \cite{peebles2023scalable} and SiT \cite{ma2024sit}, unless stated otherwise. We use ImageNet \cite{deng2009imagenet}, preprocessing images to a resolution of 256×256 following the data protocols of ADM \cite{dhariwal2021diffusion}. The images are mapped into a compressed latent representation, $\mathbf{z} \in \mathbb{R}^{32 \times 32 \times 4}$, using the Stable Diffusion VAE \cite{rombach2022high}. For model configurations, we adopt the B/2, L/2, and XL/2 architectures from DiT and SiT, using a patch size of 2. To ensure a fair comparison with DiT, SiT, and REPA, we maintain a fixed batch size of 256 during training. For adversarial alignment, we use the first two blocks of a pretrained ResNet-18 \cite{he2016deep} as the discriminator to distinguish between representations from the pretrained vision encoder and the diffusion model. To match the discriminator’s input dimensions, we apply a randomly initialized convolutional layer to adjust representation dimensions before feeding them into the network. Additional implementation details, including hyperparameters and computational resources, are provided in Appendix A.

\noindent
\textbf{Sampler.} We adopt the SDE Euler-Maruyama sampler and set the default number of function evaluations (NFE) to 250.

\noindent
\textbf{Evaluation Metrics.} We evaluate our method using 50,000 samples and report Frechet Inception Distance (FID) \cite{heusel2017gans}, sFID \cite{nash2021generating}, Inception Score (IS) \cite{salimans2016improved}, Precision (Pre.), and Recall (Rec.) \cite{kynkaanniemi2019improved}.

\noindent
\textbf{Baseline.} We compare our method against several recent diffusion-based generation approaches, each employing distinct input representations and network architectures. Specifically, we consider four categories of baselines: (1) Pixel-space diffusion models, which operate directly on image pixels, including ADM \cite{dhariwal2021diffusion}, VDM++ \cite{kingma2023understanding}, Simple Diffusion \cite{hoogeboom2023simple}, and CDM \cite{ho2022cascaded}. (2) Latent diffusion models with U-Net backbones, represented by LDM \cite{rombach2022high}. (3) Latent diffusion models incorporating transformer-U-Net hybrid architectures, such as U-ViT-H/2 \cite{bao2023all}, DiffiT \cite{hatamizadeh2024diffit}, and MDTv2-XL/2 \cite{gao2023mdtv2}. These models combine transformers with U-Net-like components, often integrating skip connections for enhanced representation propagation. (4) Latent diffusion models based purely on transformers, including MaskDiT \cite{zheng2023fast}, SD-DiT \cite{zhu2024sd}, DiT \cite{peebles2023scalable}, and SiT \cite{ma2024sit}.

\subsection{Main Results}
\label{subsec:result}
We present a comprehensive evaluation of various DiT and SiT models trained with SARA, analyzing their performance across multiple benchmarks. Additionally, we perform a system-level comparison against recent state-of-the-art diffusion models and diffusion transformers trained with REPA, demonstrating the effectiveness of our proposed alignment strategy. All models are aligned with DINOv2-B representations using $\lambda=0.5$, $\beta=0.5$, and $\gamma=0.05$. For the base model, we use the 4th-layer hidden states, while for the large and xlarge models, we use the 8th-layer hidden states. We conduct a detailed comparison under two settings: without classifier-free guidance (w/o CFG) and with classifier-free guidance (CFG). For each setting, we provide both quantitative evaluations and qualitative assessments of the generated results.

\noindent
\textbf{w/o CFG.} As shown in Figure \ref{fig:main}, under the SiT-XL/2 configuration, SARA consistently achieves lower FID scores than REPA across all training iterations. Notably, at 2M iterations, SARA attains an FID of 5.7, surpassing REPA’s performance at 4M iterations (FID = 6.1). This corresponds to a 50\% reduction in training time, significantly improving efficiency. As summarized in Table \ref{tab:wocfg}, SARA outperforms REPA across all model variants, highlighting its effectiveness in enhancing generation quality across diffusion transformers of varying scales and architectures. Figure \ref{fig:iteration} provides a qualitative comparison at different training stages, where images are generated using the same initial noise across different models. The results clearly illustrate that models trained with SARA exhibit superior progression, producing higher-quality images at every stage of training.

\noindent
\textbf{CFG.} We further evaluate SiT-XL/2 under classifier-free guidance (CFG) and quantitatively compare it against recent state-of-the-art diffusion models across multiple metrics. Without extensive hyperparameter tuning, we set the classifier-free guidance scale to $w=1.35$, consistent with REPA. As shown in Table \ref{tab:cfg}, at 200 epochs, SARA achieves performance comparable to REPA while requiring 4$\times$ fewer training epochs. Moreover, with 7$\times$ fewer epochs, SARA surpasses the original SiT-XL/2. With extended training, at 400 epochs, SiT-XL/2 trained with SARA achieves an FID of 1.75, outperforming REPA while requiring only half the training time. By further optimizing classifier-free guidance scheduling and guidance intervals, SARA reaches a state-of-the-art FID of 1.36. Figure \ref{fig:select} presents qualitative results from SiT-XL/2 trained with SARA, demonstrating its improved synthesis quality. Additional examples are provided in Appendix C.

\begin{table}
  \centering
  \resizebox{!}{0.06\textheight}{
  \begin{tabular}{@{}c@{\hspace{40pt}}c@{\hspace{40pt}}c@{\hspace{40pt}}c@{\hspace{40pt}}c}
    \toprule
    + Patch & + Struc & + Adv & FID↓ & IS↑\\
    \midrule
    $\checkmark$ &  &  & 7.9 & 122.6 \\
    \midrule
    $\checkmark$ & $\checkmark$ & & 7.3 & 131.9 \\
    \midrule
    $\checkmark$ &  & $\checkmark$ & 7.9 & 121.5 \\
    \midrule
    & $\checkmark$ & $\checkmark$ & 8.4 & 118.1 \\
    \midrule
    $\checkmark$ & $\checkmark$ & $\checkmark$ & 6.4 & 138.8 \\
    \bottomrule
  \end{tabular}}
  \caption{Ablation study for different alignment strategies}
  \label{tab:module}
\end{table}

\subsection{Ablation Study}
\label{subsec:ablation}

\textbf{Effect of Alignment Strategies.} To assess the contribution of each alignment strategy in SARA, we conduct an ablation study by systematically evaluating different combinations of patch-wise alignment (Patch), structural autocorrelation alignment (Struc), and adversarial distribution alignment (Adv). We train SiT-XL/2 models for 400K iterations, setting the coefficients to $\lambda=0.5,\beta=0.5$, and $\gamma=0.05$ when the corresponding alignment strategy is applied. As shown in Table \ref{tab:module}, using only REPA provides a strong baseline but leaves room for further improvement. Incorporating structural alignment (Struc) significantly enhances representation consistency, leading to lower FID and higher IS. The addition of adversarial alignment (Adv) further refines the representation distribution, achieving the best overall performance when all three strategies are combined. These results highlight the effectiveness of our multi-level alignment framework, demonstrating that each component contributes to improving generation quality.

\noindent
\textbf{Effect of $\beta, \gamma$.} We further examine the impact of the structural alignment coefficient $\beta$ and the adversarial alignment coefficient $\gamma$ by training SiT-XL/2 models for 400K iterations while keeping the regularization coefficient fixed at $\lambda=0.5$. As shown in Table \ref{tab:coeff}, performance improves as $\beta$ and $\gamma$ increase, reaching its optimal values at $\beta=0.5$ and $\gamma=0.05$. However, due to computational constraints, we did not explore a broader range of coefficient combinations, and finer-grained adjustments could potentially yield further improvements.

\noindent
\textbf{Effect of Pretrained Encoder.} We investigate the effect of pretrained encoders on SARA through experiments on ImageNet-256, examining how encoder type, model size, and feature extraction depth influence performance. Detailed results are presented in Appendix B.

\subsection{Discussion}
\label{subsec:discuss}
While our method demonstrates strong performance in image generation through structural and adversarial representation alignment, several limitations remain, presenting opportunities for future improvement.

\noindent
\textbf{Dependence on Pretrained Visual Encoders.} Our approach relies on pretrained visual encoders, which may introduce domain gaps when applied to datasets with different distributions. Future work could explore adaptive representation learning to reduce dependence on external models.

\noindent
\textbf{Sensitivity to Hyperparameters.} Balancing multiple loss terms requires careful tuning, and suboptimal choices can degrade performance. Future research could investigate automated weight selection strategies to optimize training stability and efficiency.

\noindent
\textbf{Spatial Coherence in High-Resolution Images.} While SARA improves representation consistency, high-resolution generation may still suffer from local artifacts or texture inconsistencies. Multi-scale alignment could further enhance coherence across resolutions.

\noindent
\textbf{Exploring Alternative Data Domains.} While our work focuses on latent diffusion for image generation, extending alignment strategies to pixel-space diffusion models or other modalities, such as video and 3D synthesis, presents an exciting research direction.

\begin{table}
  \centering
  \resizebox{!}{0.05\textheight}{
  \begin{tabular}{@{}c@{\hspace{50pt}}c@{\hspace{50pt}}c@{\hspace{50pt}}c}
    \toprule
    $\beta$ & $\gamma$ & FID↓ & IS↑ \\
    \midrule
    0.1 & 0.01 & 7.9 & 123.1 \\
    0.1 & 0.05 & 7.7 & 125.6 \\
    0.5 & 0.01 & 7.2 & 130.4 \\
    0.5 & 0.05 & 6.4 & 138.8 \\
    \bottomrule
  \end{tabular}}
  \caption{Ablation study for $\beta$ and $\gamma$}
  \label{tab:coeff}
\end{table}

\section{Conclusions}
\label{sec:conclusions}

In this work, we introduce SARA, a representation alignment framework designed to enhance diffusion transformers by extending the REPA method with structured autocorrelation and adversarial representation alignment. Extensive experiments demonstrate that our approach improves the consistency between pretrained visual encoders and diffusion transformers, leading to greater training efficiency and higher image quality. Our work paves the way for further advancements in diffusion-based models. Future research will explore adaptive alignment strategies, automated weight selection, and multimodal extensions to further advance high-quality vision generation.

\clearpage

{
     \small
     \bibliographystyle{ieeenat_fullname}

}
\vfill

\appendix


\begin{table*}[ht]
    \begin{tabular}{@{}lcccccc}
    \toprule
    & Figure 2 & Table 1a (SiT-B) & Table 1a (SiT-L) & Table 1a (SiT-XL) & Table 1b \\
    \midrule
    \textbf{Architecture} & & & & & \\
    Input dim. & $32\times32\times4$ & $32\times32\times4$ & $32\times32\times4$ & $32\times32\times4$ & $32\times32\times4$ \\
    Layers & 28 & 12 & 24 & 28 & 28 \\
    Hidden dim. & 1152 & 768 & 1024 & 1152 & 1152 \\
    Heads & 16 & 12 & 16 & 16 & 16 \\
    \midrule
    \textbf{SARA} & & & & & \\
    $\lambda$ & 0.5& 0.5 & 0.5 & 0.5 & 0.5 \\
    $\beta$ & 0.5 & 0.5 & 0.5 & 0.5 & 0.5 \\
    $\gamma$ & 0.05& 0.05 & 0.05 & 0.05 & 0.05 \\
    Alignment depth & 8 & 4 & 8 & 8 & 8 \\
    $sim(\cdot,\cdot)$ & cos. sim. & cos. sim. & cos. sim. & cos. sim. & cos. sim. \\
    Encoder $f(x)$ & DINOv2-B & DINOv2-B & DINOv2-B & DINOv2-B & DINOv2-B \\
    \midrule
    \textbf{Optimization} & & & & & \\
    Training iteration & 400K & 400K & 400K & 2M & 2M \\
    Batch size & 256 & 256 & 256 & 256 & 256 \\
    Optimizer & AdamW & AdamW & AdamW & AdamW & AdamW \\
    lr & 0.0001 & 0.0001 & 010001 & 010001 & 010001 \\
    $(\beta_1,\beta_2)$ & (0.9, 0.999)& (0.9, 0.999) & (0.9, 0.999) & (0.9, 0.999) & (0.9, 0.999) \\
    \midrule
    \textbf{Interpolants} & & & & & \\
    $\alpha_t$ & $1-t$ & $1-t$ & $1-t$ & $1-t$ & $1-t$ \\
    $\sigma_t$ & $t$ & $t$ & $t$ & $t$ & $t$ \\
    $w_t$ &  $\sigma_t$ &  $\sigma_t$ &  $\sigma_t$ &  $\sigma_t$ &  $\sigma_t$ \\
    Training objective & v-prediction & v-prediction & v-prediction & v-prediction & v-prediction \\
    Sampler & Euler-Maruyama & Euler-Maruyama & Euler-Maruyama & Euler-Maruyama & Euler-Maruyama \\
    Sampling steps & 250 & 250 & 250 & 250 & 250 \\
    Guidance & - & - & - & - & 1.35 \\ 
    \bottomrule
    \end{tabular}
    \caption{Hyperparameter setup}
    \label{tab:hyper}
\end{table*}

\section{Hyperparameters and more implementation details}

\begin{table*}[ht]
    \renewcommand{\arraystretch}{1.1}
    \begin{tabular}{@{}lc@{\hspace{50pt}}c@{\hspace{50pt}}c@{\hspace{50pt}}c@{\hspace{50pt}}c@{\hspace{50pt}}cc}
    \toprule
    Target Repr. & Depth &  FID↓ & sFID↓ & IS↑ & Pre.↑ & Rec.↑ \\
    \midrule
    SiT-L/2 & 8 & 18.8 & 5.29 & 72.9 & 0.64 & 0.64 \\
    \midrule
    MAE-L+REPA & 8 & 12.5 & 4.89 & 90.7 & 0.68 & 0.63 \\
    \textbf{MAE-L+SARA} & 8 & 11.5 & 4.97 & 102.3 & 0.69 & 0.64 \\
    MoCov3-L+REPA & 8 & 11.9 & 5.06 & 92.2 & 0.68 & 0.64 \\
    \textbf{MoCov3-L+SARA} & 8 & 11.1 & 5.02 & 98.8 & 0.69 & 0.64 \\
    CLIP-L+REPA & 8 & 11.0 & 5.25 & 107.0 & 0.69 & 0.64\\
    \textbf{CLIP-L+SARA} & 8 & 9.8 & 5.15 & 110.8 & 0.70 & 0.64\\
    \midrule
    DINOv2-B+REPA & 8 & 9.7 & 5.13 & 107.5 & 0.69 & 0.64 \\
    \textbf{DINOv2-B+SARA} & 8 & 8.5 & 5.24 & 119.3 & 0.69 & 0.66 \\
    DINOv2-L+REPA & 8 & 10.0 & 5.09 & 106.6 & 0.68 & 0.65\\
    \textbf{DINOv2-L+SARA} & 8 & 8.4 & 5.18 & 121.6 & 0.69 & 0.66 \\
    DINOv2-g+REPA & 8 & 9.8 & 5.22 & 108.9 & 0.69 & 0.64 \\
    \textbf{DINOv2-g+SARA} & 8 & 8.6 & 5.22 & 119.7 & 0.69 & 0.66 \\
    \midrule
    DINOv2-B+SARA & 6 & 8.7 & 5.33 & 116.4 & 0.69 & 0.65 \\
    DINOv2-B+SARA & 8 & 8.5 & 5.24 & 119.3 & 0.69 & 0.66 \\
    DINOv2-B+SARA & 10 & 9.1 & 5.34 & 112.3 & 0.69 & 0.65 \\
    DINOv2-B+SARA & 12 & 9.9 & 5.15 & 110.8 & 0.70 & 0.64 \\
    DINOv2-B+SARA & 14 & 10.4 & 5.14 & 107.6 & 0.69 & 0.64 \\
    DINOv2-B+SARA & 16 & 11.2 & 5.17 & 102.4 & 0.69 & 0.64 \\
    \bottomrule
    \end{tabular}
    \caption{Ablation study of pretrained encoders on ImageNet-256. All models are SiT-L/2 trained for 400K iterations. All metrics are measured with the SDE Euler-Maruyama sampler with NFE=250 and without classifier-free guidance. We fix $\lambda=0.5,\beta=0.5,\gamma=0.05$ here. ↓ and ↑ indicate whether lower or higher values are better, respectively.}
    \label{tab:enc}
\end{table*}

\noindent
\textbf{Further implementation details.} To ensure a fair comparison, our experimental setup is nearly identical to REPA. Specifically, we adopt the same architecture as DiT and SiT throughout all experiments. We use AdamW as the optimizer with a learning rate of 1e-4, $(\beta_1,\beta_2)=(0.9,0.999)$, and no weight decay. To accelerate training, we employ mixed-precision (fp16) training with gradient clipping and precompute compressed latent vectors from raw pixels using Stable Diffusion VAE, which are then used throughout training. For the MLP used in projection, we adopt a three-layer MLP with SiLU activation. A detailed hyperparameter configuration is provided in Table \ref{tab:hyper}.

\noindent
\textbf{Distriminator details.} For adversarial alignment training, we employ a lightweight discriminator to minimize computational overhead. Specifically, we use a pretrained ResNet-18, removing its final two resdual blocks to reduce complexity. Additionally, we modify the first convolutional layer to a $1\times1$ convolution, allowing it to transform transformer-based representations into a suitable input dimension for the discriminator. During training, the discriminator distinguishes between features extracted by the diffusion model (negative samples) and features from the pretrained vision encoder (positive samples). Since the discriminator is not trained from scratch, we adopt an asymmetric training strategy—for every five updates to the diffusion model, the discriminator is updated only once. This strategy stabilizes adversarial training and prevents the generator from diverging in the early stages due to an overly strong discriminator.

\noindent
\section{Effect of diffirent pretrained encoders} As shown in Table \ref{tab:enc}, SARA consistently improves generation quality across various pretrained encoders, outperforming both the original model (SiT) and REPA in terms of FID scores. This demonstrates that our approach effectively leverages pretrained representations for enhanced synthesis. Next, we evaluate the impact of encoder size by comparing different DINOv2 variants. Consistent with observations in REPA, we find that increasing the encoder size cannot leads to marginal performance improvements. This suggests that SARA primarily benefits from the structural alignment of features rather than the absolute model capacity. Finally, we examine the importance of representation extraction depth and observe that aligning only the early-layer representations of the pretrained encoder is sufficient to achieve strong performance. This finding indicates that lower-level representations contain enough information for effective alignment, reducing the need for deeper feature supervision. These results highlight the robustness of SARA across different pretrained encoder configurations and suggest that carefully selecting the alignment depth can improve efficiency without compromising quality.

\section{More qualitative results}

\begin{figure*}
  \centering
  \includegraphics[width=\textwidth]{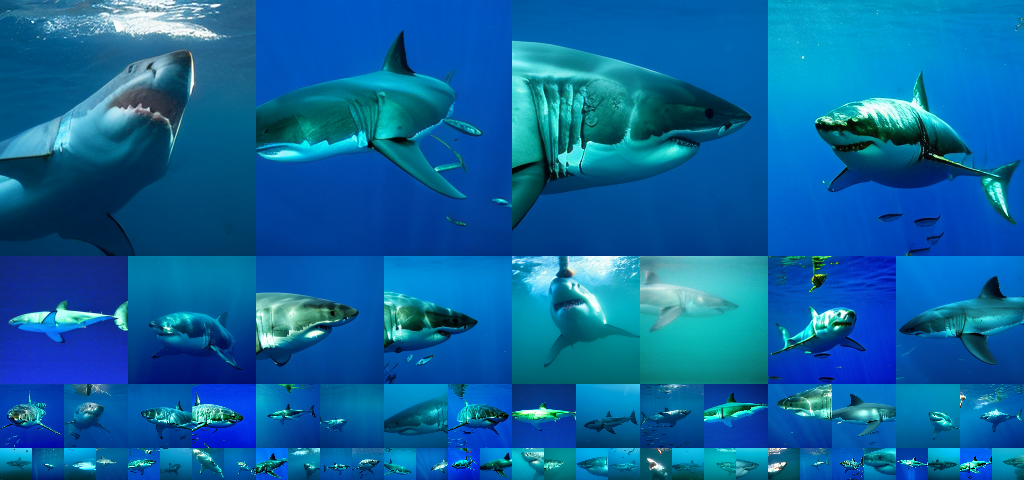}
  \caption{Uncurated generation results of SiT-XL/2+REPA. We use classifier-free guidance with $w=4.0$. Class label=“great white shark”(2).}
\end{figure*}

\begin{figure*}
  \centering
  \includegraphics[width=\textwidth]{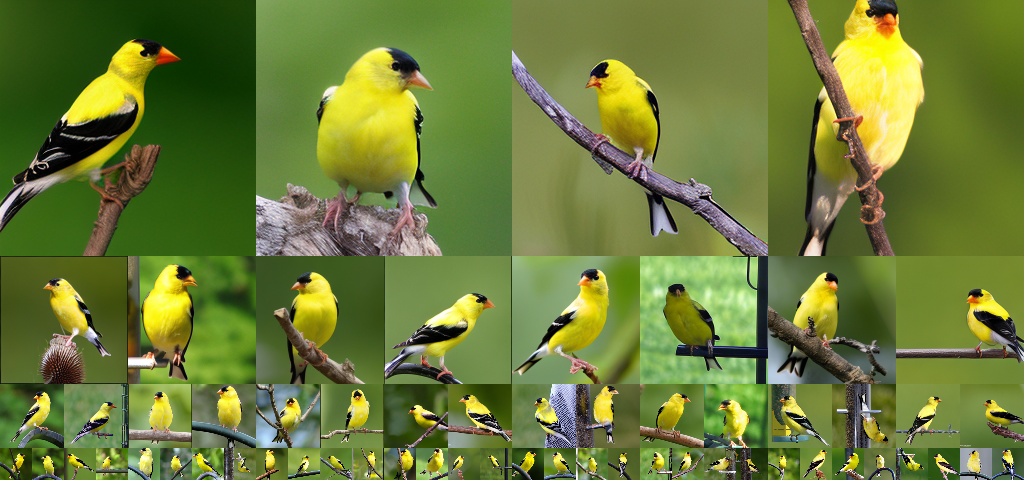}
  \caption{Uncurated generation results of SiT-XL/2+REPA. We use classifier-free guidance with $w=4.0$. Class label=“goldfinch”(11).}
\end{figure*}

\begin{figure*}
  \centering
  \includegraphics[width=\textwidth]{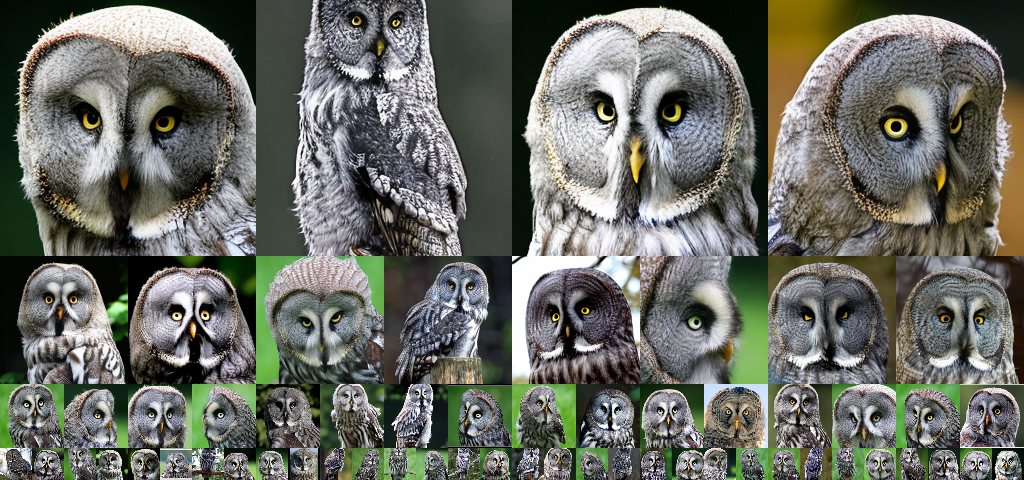}
  \caption{Uncurated generation results of SiT-XL/2+REPA. We use classifier-free guidance with $w=4.0$. Class label=“great grey owl”(24).}
\end{figure*}

\begin{figure*}
  \centering
  \includegraphics[width=\textwidth]{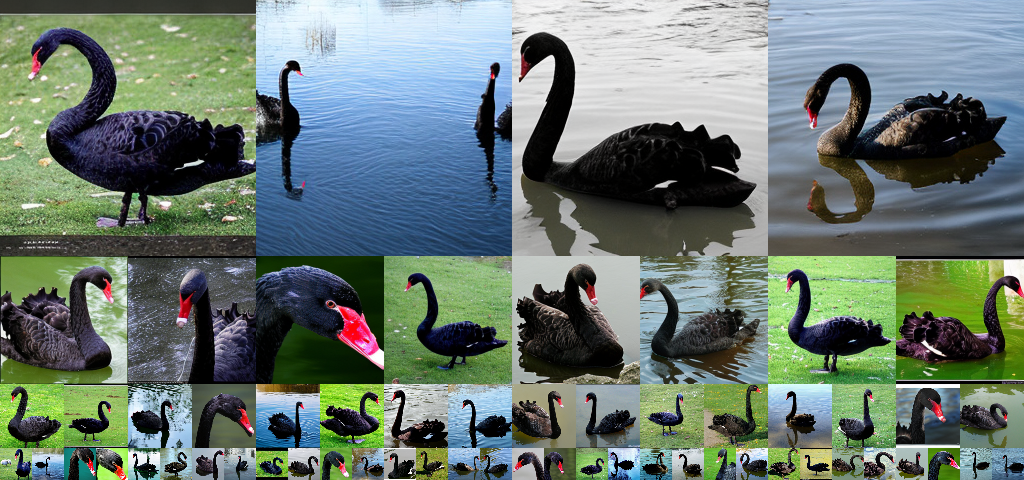}
  \caption{Uncurated generation results of SiT-XL/2+REPA. We use classifier-free guidance with $w=4.0$. Class label=“black swan”(100).}
\end{figure*}

\begin{figure*}
  \centering
  \includegraphics[width=\textwidth]{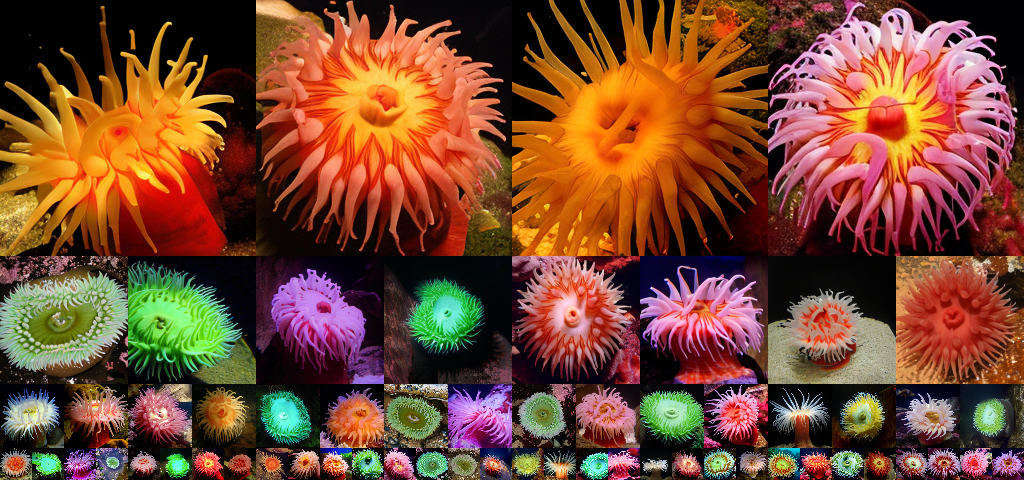}
  \caption{Uncurated generation results of SiT-XL/2+REPA. We use classifier-free guidance with $w=4.0$. Class label=“sea anemone”(108).}
\end{figure*}

\begin{figure*}
  \centering
  \includegraphics[width=\textwidth]{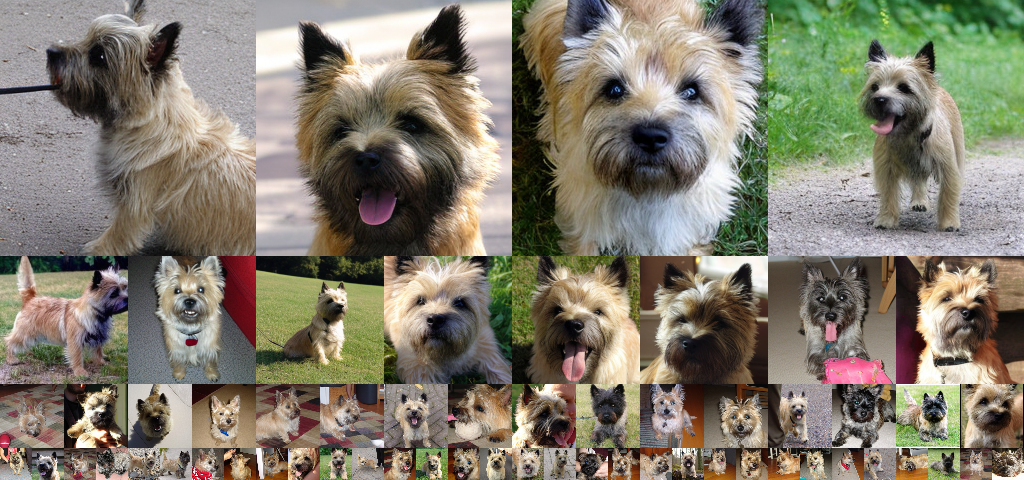}
  \caption{Uncurated generation results of SiT-XL/2+REPA. We use classifier-free guidance with $w=4.0$. Class label=“cairn”(192).}
\end{figure*}

\begin{figure*}
  \centering
  \includegraphics[width=\textwidth]{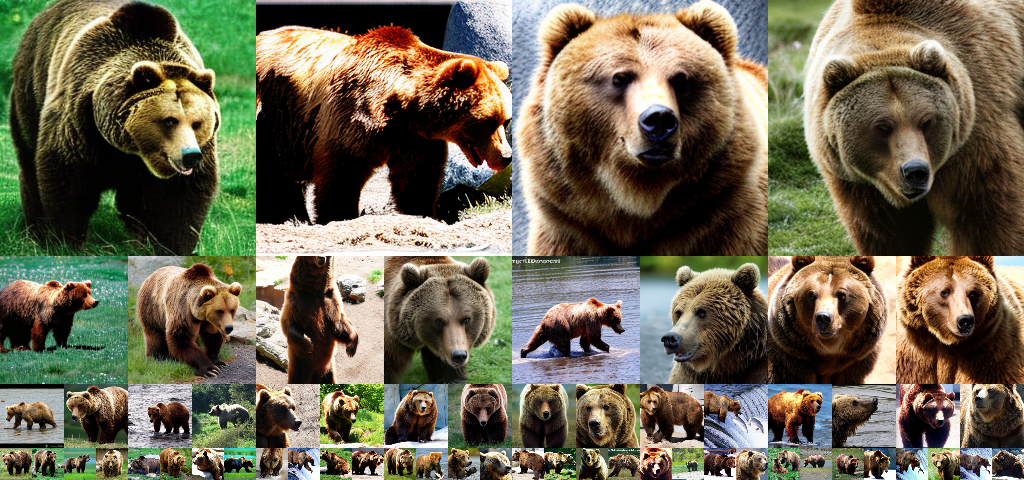}
  \caption{Uncurated generation results of SiT-XL/2+REPA. We use classifier-free guidance with $w=4.0$. Class label=“brown bear”(294).}
\end{figure*}

\begin{figure*}
  \centering
  \includegraphics[width=\textwidth]{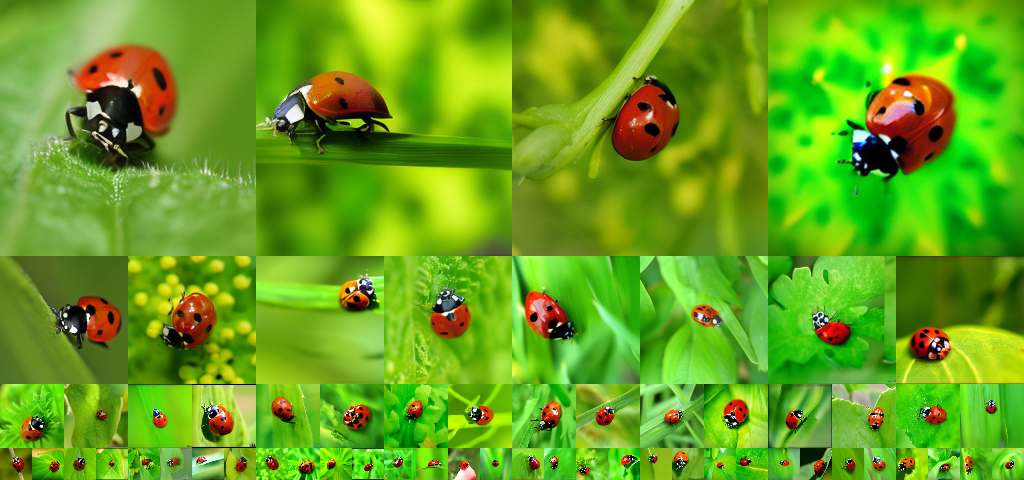}
  \caption{Uncurated generation results of SiT-XL/2+REPA. We use classifier-free guidance with $w=4.0$. Class label=“ladybug”(301).}
\end{figure*}

\begin{figure*}
  \centering
  \includegraphics[width=\textwidth]{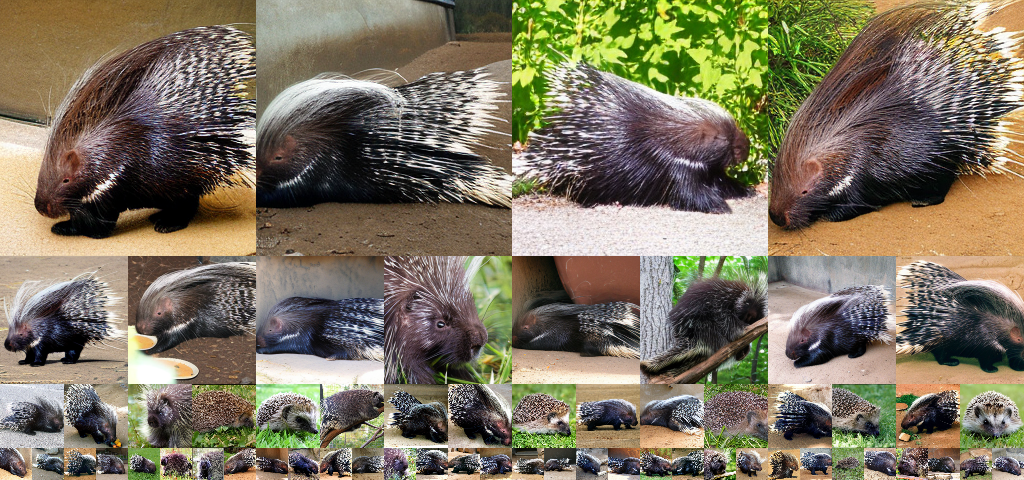}
  \caption{Uncurated generation results of SiT-XL/2+REPA. We use classifier-free guidance with $w=4.0$. Class label=“porcupine”(334).}
\end{figure*}

\begin{figure*}
  \centering
  \includegraphics[width=\textwidth]{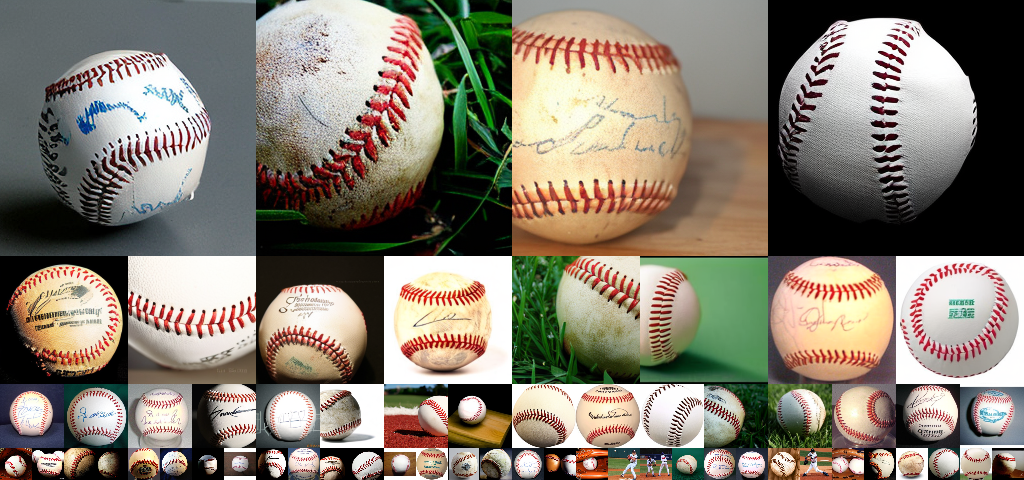}
  \caption{Uncurated generation results of SiT-XL/2+REPA. We use classifier-free guidance with $w=4.0$. Class label=“baseball”(429).}
\end{figure*}

\begin{figure*}
  \centering
  \includegraphics[width=\textwidth]{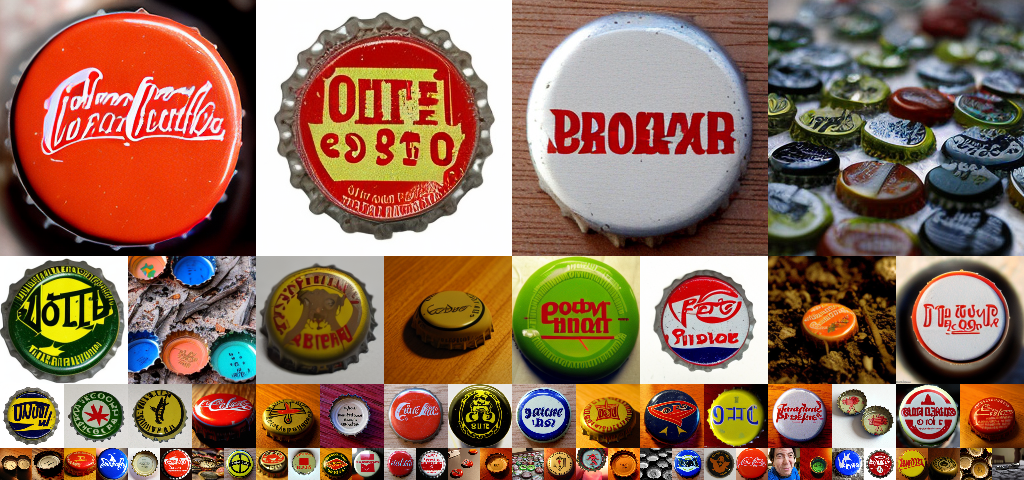}
  \caption{Uncurated generation results of SiT-XL/2+REPA. We use classifier-free guidance with $w=4.0$. Class label=“bottlecap”(455).}
\end{figure*}

\begin{figure*}
  \centering
  \includegraphics[width=\textwidth]{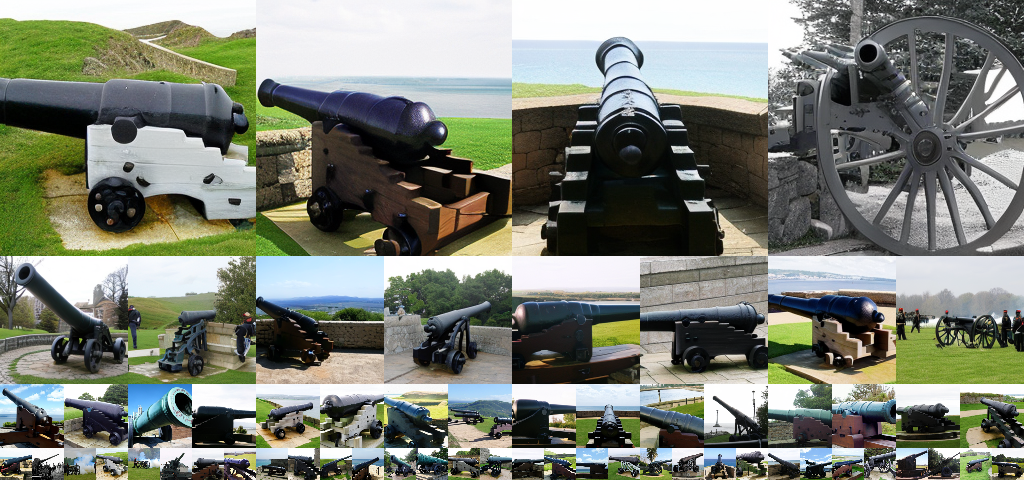}
  \caption{Uncurated generation results of SiT-XL/2+REPA. We use classifier-free guidance with $w=4.0$. Class label=“cannon”(471).}
\end{figure*}

\begin{figure*}
  \centering
  \includegraphics[width=\textwidth]{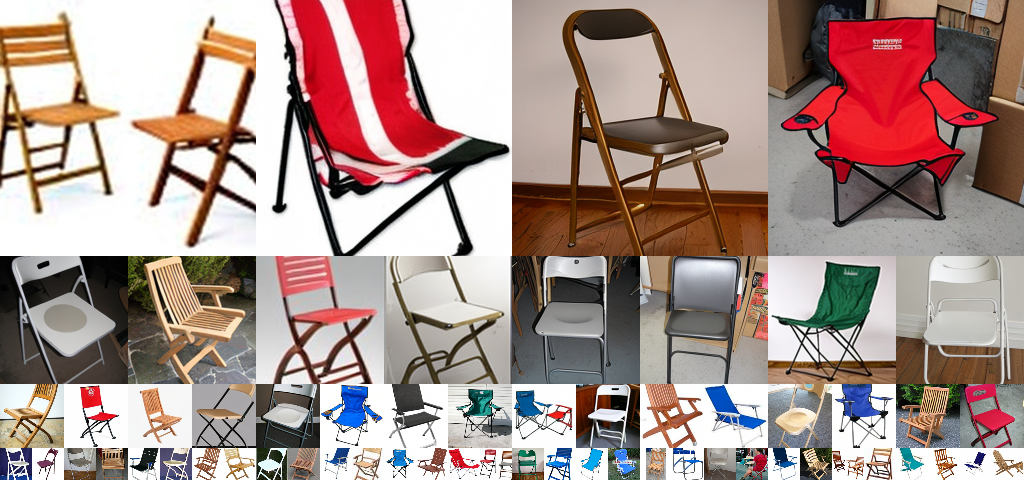}
  \caption{Uncurated generation results of SiT-XL/2+REPA. We use classifier-free guidance with $w=4.0$. Class label=“folding chair”(559).}
\end{figure*}

\begin{figure*}
  \centering
  \includegraphics[width=\textwidth]{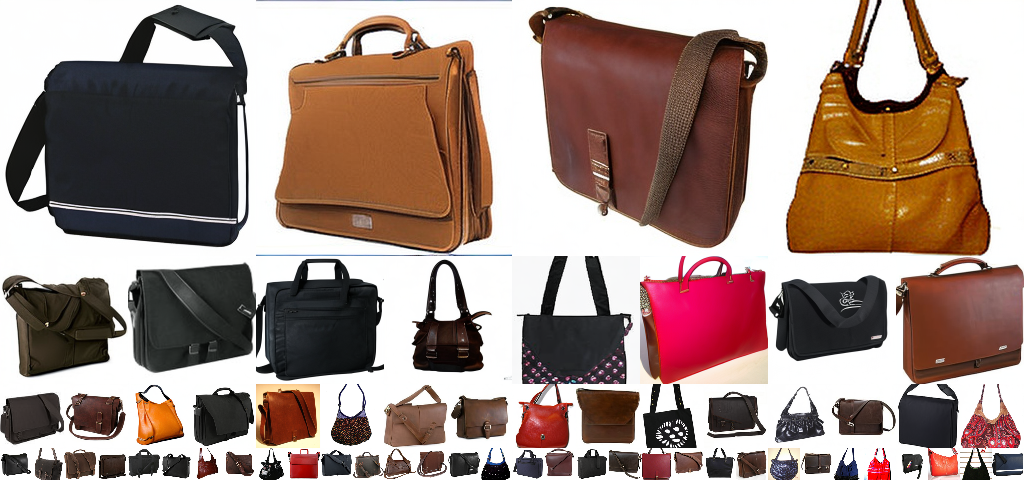}
  \caption{Uncurated generation results of SiT-XL/2+REPA. We use classifier-free guidance with $w=4.0$. Class label=“mailbag”(636).}
\end{figure*}

\begin{figure*}
  \centering
  \includegraphics[width=\textwidth]{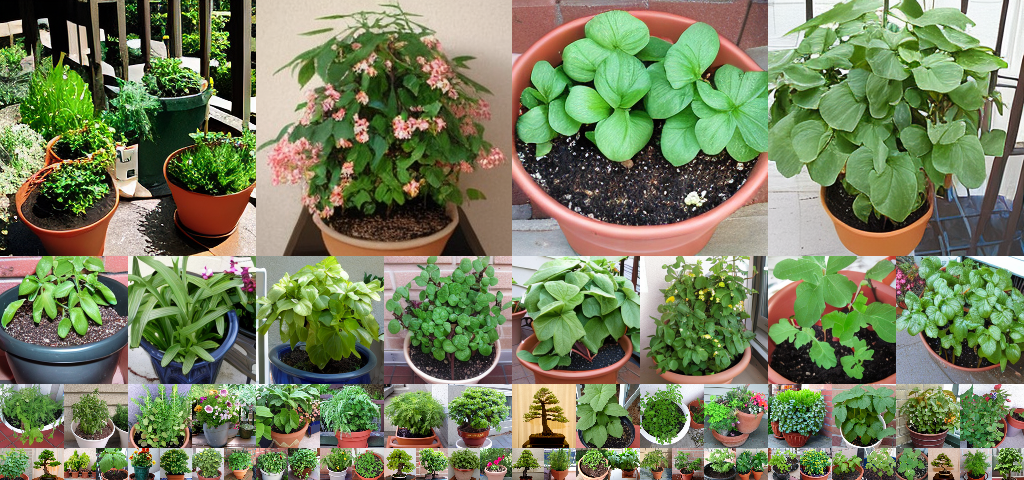}
  \caption{Uncurated generation results of SiT-XL/2+REPA. We use classifier-free guidance with $w=4.0$. Class label=“pot”(738).}
\end{figure*}

\begin{figure*}
  \centering
  \includegraphics[width=\textwidth]{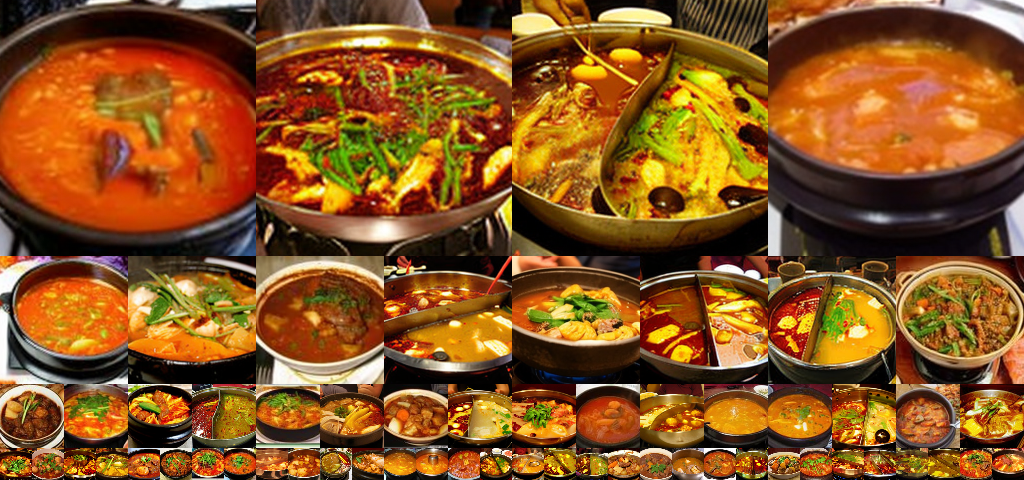}
  \caption{Uncurated generation results of SiT-XL/2+REPA. We use classifier-free guidance with $w=4.0$. Class label=“hot pot”(926).}
\end{figure*}

\end{document}